\documentclass[sigconf]{acmart}

\AtBeginDocument{%
  \providecommand\BibTeX{{%
    \normalfont B\kern-0.5em{\scshape i\kern-0.25em b}\kern-0.8em\TeX}}}

\setcopyright{acmcopyright}
\copyrightyear{2018}
\acmYear{2018}
\acmDOI{10.1145/1122445.1122456}

\acmConference[KDD '20]{KDD 2020: Workshop on Applied Data Science for Healthcare}{August 24, 2020}{San Diego, CA}
\acmBooktitle{KDD 2020: Workshop on Applied Data Science for Healthcare, August 24 2020, San Diego, CA}
\acmPrice{15.00}
\acmISBN{978-1-4503-XXXX-X/18/06}

\begin{document}

\title{MCU-Net: A framework towards uncertainty representations for decision support system patient referrals in healthcare contexts}

\author{Nabeel Seedat}
\authornotemark[1]
\email{seedatnabeel@gmail.com}
\affiliation{%
  \institution{Cornell University \& Shutterstock}
  \city{New York}
  \state{NY}
}

\renewcommand{\shortauthors}{N.Seedat}

\begin{abstract}
Incorporating a human-in-the-loop system when deploying automated decision support is critical in healthcare contexts to create trust, as well as provide reliable performance on a patient-to-patient basis. Deep learning methods while having high performance, do not allow for this patient-centered approach due to the lack of uncertainty representation. 

Thus, we present a framework of uncertainty representation evaluated for medical image segmentation, using MCU-Net which combines a U-Net with Monte Carlo Dropout, evaluated with four different uncertainty metrics. The framework augments this by adding a human-in-the-loop aspect based on an uncertainty threshold for automated referral of uncertain cases to a medical professional.  

We demonstrate that MCU-Net combined with epistemic uncertainty and an uncertainty threshold tuned for this application maximizes automated performance on an individual patient level, yet refers truly uncertain cases. This is a step towards uncertainty representations when deploying machine learning based decision support in healthcare settings. 
\end{abstract}



\begin{teaserfigure}
\begin{centering}
  \includegraphics[scale=0.45,trim={0.1cm 0.1cm 0.1cm 0.1cm},clip]{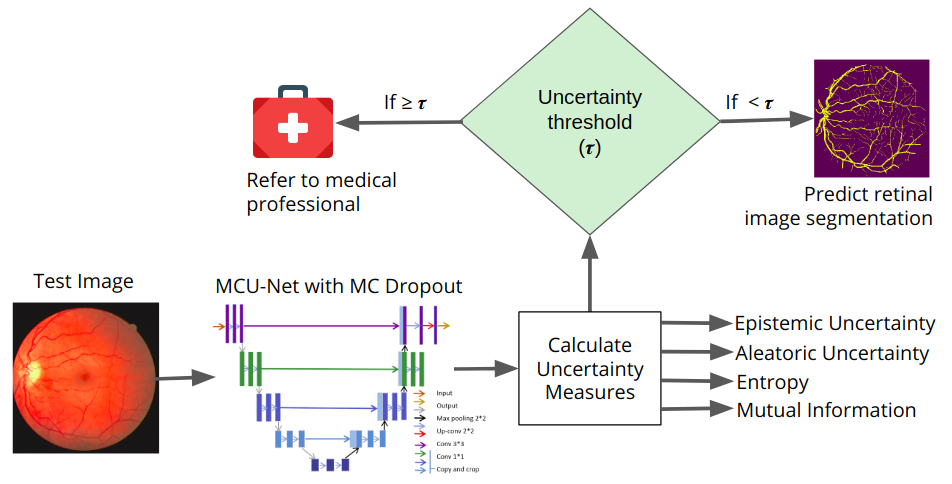}
  \caption{Our proposed framework for uncertainty representation in biomedical image segmentation. It incorporates a medical-professional-in-the-loop based on uncertainty}
  \Description{Enjoying the baseball game from the third-base
  seats. Ichiro Suzuki preparing to bat.}
  \label{fig:teaser}
  \end{centering}
\end{teaserfigure}

\maketitle

\section{Introduction}
Deep learning has enabled outstanding performance in many computer vision tasks, including medical image analysis \cite{medical}. However, for applications in a critical domain like healthcare, it is imperative that neural networks provide estimates of uncertainty \cite{seedat19}. Unfortunately, current off-the-shelf models lack this capability~\cite{gal2015dropout,maddoxfast,Pearce2018UncertaintyIN,sri, seedat19}. Furthermore, the softmax outputs are poor measures for confidence in the prediction, as they often result in overconfident predictions due to miscalibration~\cite{nguyen2015deep,Kendall2017,seedat19}.

Bayesian Neural Networks (BNNs) offer a principled approach to uncertainty estimation in neural networks, by providing a probabilistic interpretation of predictive distributions \cite{seedat19}. While Bayesian methods typically present a computational intractability, Monte Carlo Dropout (MCD) \cite{gal2015dropout} has been shown to address the computational issue by formulating conventional dropout as an equivalent to Bayesian variational inference.  

That being said, whilst many works usually focus on inference, they often use different uncertainty metrics. Thus, it is unclear when using BNNs which uncertainty metric is the most appropriate for different applications \cite{seedat19}. Finally, while most models are evaluated based on generalization to a test set, what makes applying such models to healthcare unique is that individual patient-by-patient performance is more important rather than aggregated cohort/test-set results.

Hence, it is critical that models should convey uncertainty in decisions, so that individual highly uncertain cases may be flagged for referral to a medical professional in an automated manner. This human-in-the-loop aspect would provide transparent and safer patient-centered care, as well as allow for optimal allocation of constrained hospital resources.\\\\

This paper makes the following contributions:\\
(1)  Investigate uncertainty representations in medical image segmentation using our proposed model called MCU-Net (\textbf{M}onte \textbf{C}arlo \textbf{U}-Net), which combines a U-Net with Monte-Carlo Dropout for uncertainty representation. \\
(2) Evaluate and compare the efficacy of different uncertainty metrics for medical image segmentation.\\
(3) We propose an automated framework whereby uncertainty representations enable a human-in-the-loop system in the healthcare context. Specifically to flag uncertain cases for referral to medical professionals, ensuring safer and transparent outcomes under uncertainty.

\section{Our Method: MCU-Net with Uncertainty Thresholds}
We propose a framework, illustrated in Figure 1, for uncertainty representation in healthcare settings, which enables 
a human-in-the-loop referral of cases. The framework is studied on medical image segmentation, however the framework can be generalized to other healthcare domains or medical imaging tasks. 

The framework consists of two components: Firstly, a proposed model called \textbf{M}onte \textbf{C}arlo \textbf{U}-Net (MCU-Net) which incorporates uncertainty in image segmentation. Secondly, an evaluation of uncertainty metrics leading to a principled uncertainty threshold ($\tau$) that would allow for automated flagging and referral of cases to medical professionals.

\subsection{\textbf{M}onte \textbf{C}arlo \textbf{U}-Net (MCU-Net) }
We present MCU-Net, which incorporates an uncertainty representation into the task of medical image segmentation. The method combines a U-Net widely used for biomedical image segmentation \cite{ronneberger2015u}, with Monte Carlo Dropout (MCD) \cite{gal2015dropout}. 

By this we mean applying the U-Net to perform image segmentation, whilst MCD is then used for approximate Bayesian inference. This involves performing N Monte Carlo samples, which is achieved by performing N forward passes through the U-Net (i.e. infer $y|x$ N times). At each iteration, we sample a different set of network units to drop out. This generates stochastic predictions, which are interpreted as samples from a probabilistic distribution \cite{gal2015dropout}. 

Thereafter, the uncertainty in the segmentation predictions is captured by evaluating four different uncertainty metrics on the aforementioned probabilistic samples.\\

The four metrics are:
\begin{itemize}
    \item \textit{Aleatoric Uncertainty:} which captures the inherent noise (stochasticity) in the data \cite{Kendall2017,sri,Gal2017} and is calculated as per \cite{kwon2018uncertainty}: $\frac{1}{T}\sum_{t=1}^{T}diag(\hat{p_{t}})-\hat{p_{t}}\hat{p_{t}}^{T}$,  where, $\hat{p_{t}} = $ softmax $(f_{w_{t}}(x^{*})$).
\item \textit{Epistemic Uncertainty:} which
is the inherent model uncertainty \cite{Kendall2017,sri,Gal2017}, where data that is different from training should have a higher epistemic uncertainty. It is calculated as per \cite{kwon2018uncertainty}: $\frac{1}{T}\sum_{t=1}^{T}(\hat{p_{t}}-\bar{p_{t}})(\hat{p_{t}}-\bar{p_{t}})^{T} $ where  $ \bar{p_{t}}$  = $\frac{1}{T}\sum_{t=1}^{T}\hat{p_{t}}$.
\item \textit{Predictive Entropy:} where a higher entropy corresponds to a greater amount of uncertainty \cite{mackay2003information}. It is calculated as $ H=-\sum_{y \in Y}^{} P(y|x)logP(y|x)$, where $P(y|x)$ is the softmax output.
\item \textit{Mutual Information:}
 is the information gain related to the model parameters for the dataset if we see a label $y$ for an input $x$. It is the predictive entropy minus expected entropy given by: 
$ MI = H[P(y | x,D)]-\mathbb{E}_{p(w|D)}H[P(y|x,w)]$
\end{itemize}

\subsection{Uncertainty thresholds}
As illustrated in Figure 1, we then aim to ascertain the optimal uncertainty threshold ($\tau$). This threshold will differ based on the application. However, we present a preliminary analysis  using the medical imaging case study. The segmentation cases that exceed the uncertainty threshold ($\tau$) are then flagged for referral to a medical-professional-in-the-loop. 

For real-world application in a healthcare context, we propose that the optimal $\tau$ is quantified as maximizing the model performance on individual cases/patients (by only making predictions on cases with good certainty), whilst not referring too many cases such that the benefit of automated diagnosis is mitigated. i.e. mitigated where the model only evaluates simpler, highly certain cases, whilst referring too many cases such that it provides no reduction in clinical workload. This trade-off is detailed in the experimental evaluation.

\begin{figure*}[h]
    \centering
    \includegraphics[scale=0.33]{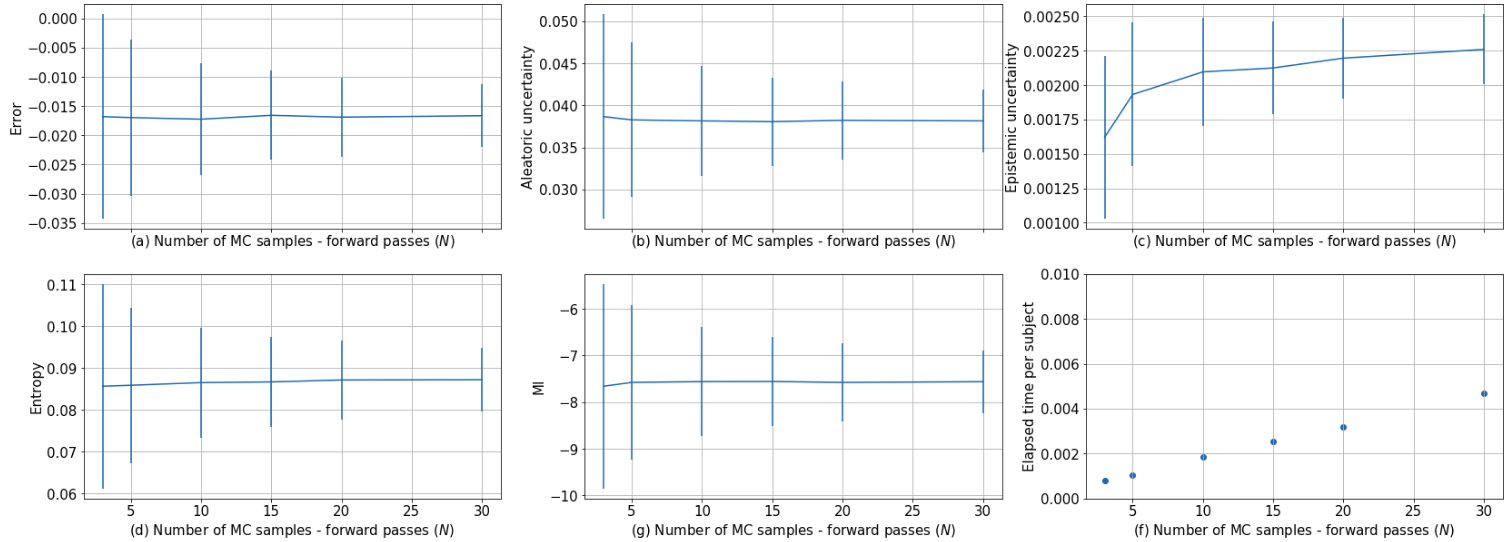}
    \caption{Uncertainty measures for different numbers of MC samples from N=1-30}
    \label{fig:wdist}
\end{figure*}

\section{Experimental evaluation}
We carry out a preliminary evaluation of the aforementioned framework presented in Figure 1 using the Digital Retinal Images for Vessel Extraction (DRIVE) dataset \cite{drive}, which can be found at https://github.com/seedatnabeel/Uncertainty-Decision-Support-Healthcare. The dataset contains 40 labelled images (20 train and 20 test) to evaluate segmentation of blood vessels in retinal images. 

It is imperative that uncertainty is incorporated in this process of vessel segmentation as the results are used for detection and analysis of vessels in the diagnosis, screening and treatment of diseases such as diabetes, hypertension and arteriosclerosis \cite{drive}.

The experimental evaluation involves: (1) the evaluation of MCU-Net on the task of blood vessel segmentation using the uncertainty metrics and (2) determining the optimal uncertainty threshold ($\tau$). We perform evaluation with a standard U-Net \cite{ronneberger2015u} initialized using He Normal Initialization \cite{he2015delving}.  

Approximate Bayesian inference is performed using Monte Carlo dropout, with dropout probability of 0.25. Finally, given the small dataset size, we augment the data by training the network on 1000 random patches from the training set and evaluate using 100 random patches from the test set.
	
\subsection{MCU-Net evaluation}
We assess MCU-Net using the networks predictive probabilities evaluated using the four aforementioned uncertainty metrics, as well as, evaluating the overall error and execution time. The mean and standard deviation are reported for these measures. We quantify the impact of different numbers of Monte Carlo samples, for N ranging from one to thirty stochastic forward passes. 

The results are presented in Figure 2 and it is evident that as the number of MC samples increases, the variance in the uncertainty metrics decreases. That being said, epistemic uncertainty (model uncertainty) increases till a knee-point of 20 MC samples. 

Since 20 samples indicates a stability point in the metrics with the lowest execution time, we use it to analyze the performance for retinal vessel segmentation. Additionally, we evaluate which uncertainty metric is most useful in conveying the representation of uncertainty. The segmentation results for the different uncertainty metrics is shown in Figure 3.

\begin{figure}[h]
    \centering
    \includegraphics[scale=0.35]{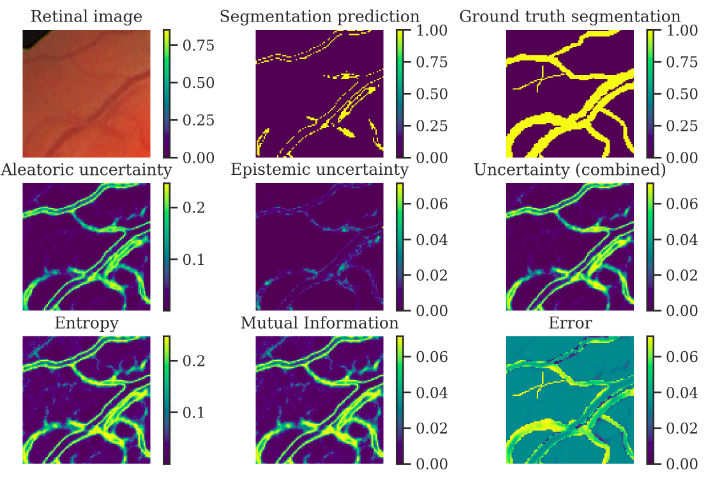}
    \caption{Segmentation results for the retinal images. An example of the original retinal image, predicted segmentation, ground truth segmentation and different uncertainty metrics is illustrated}
    \label{fig:plots}
\end{figure}

As illustrated in Figure 3 (and for other examples not shown), the model has difficulty segmenting the narrower branches of the vessels. Aleatoric uncertainty and entropy give similar performance, and likewise for  mutual information and the combination of uncertainty (aleatoric + epistemic). In particular, these methods convey high uncertainty for most of the segmented region.

This is contrasted with epistemic uncertainty which provides a finer grained representation of the areas where the model has difficulty on the narrower vessels. Hence, suggesting that epistemic uncertainty is the most representative uncertainty metric.

\begin{figure}[t]
    \centering
    \includegraphics[scale=0.4]{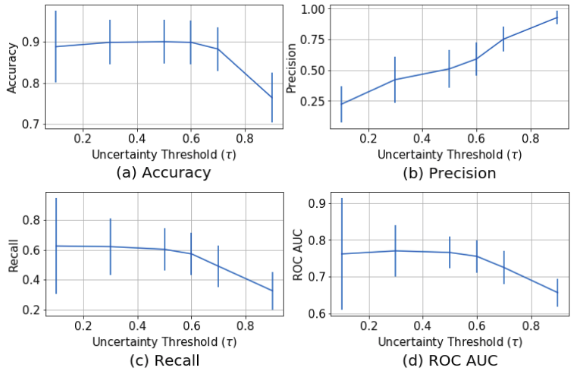}
    \caption{Performance metrics for different values of the uncertainty threshold $\tau$. As $\tau$ increases the model is less cautious and fewer cases are referred. }
    \label{fig:plots}
\end{figure}

\subsection{Optimal uncertainty threshold ($\tau$)}
The uncertainty threshold $\tau$ is defined as the proportion of the maximum uncertainty (per case). Referrals then use this value of $\tau$ (i.e. proportion), such that cases that exceed this proportion (uncertainty threshold) are referred to a clinician-in-the-loop. We evaluate values of $\tau$ between the range of 0.1-0.9. Thereafter, we mimic the real healthcare workflow of referring uncertain cases for a second opinion to a medical professional. This is achieved by removing those cases from the model's test set that have uncertainty greater than the threshold. 

Model performance on the remaining cases is assessed based on the accuracy, precision, recall and AUROC for each value of $\tau$. It is expected that as the uncertainty threshold ($\tau$) increases, that the model is less cautious in decision making, thereby making predictions on more cases despite the increase in uncertainty. This means that for greater values of $\tau$, performance will likely decrease, as fewer cases are referred to the medical professional, even under high uncertainty.

However, we wish to balance high performance (accuracy, precision, recall, area under ROC) with having a higher threshold ($\tau$), in order that more samples are evaluated autonomously rather than being referred. The results of this experimentation is shown in Figure 4. The results indicate that with increasing $\tau$, the accuracy, AUROC and recall is steady till $\tau$ of 0.6 and thereafter the performance metrics decrease as more predictions are made when the model is uncertain. 
 
 The appropriate uncertainty threshold would naturally be task specific, as well as take into account clinical guidance.  In this specific segmentation task the performance metrics are calculated on a per-pixel level. Hence, there is tolerance of marginally lower precision in favor of higher recall. 
 
 Thus, we propose a threshold ($\tau$) of 0.6 for this preliminary study to best satisfy the performance with certainty vs automation trade-off. This chosen uncertainty threshold would result in only medium-highly uncertain cases being referred to a medical-professional-in-the-loop. Whilst, on the cases that are retained there is confidence of high performance given the certainty scores.  
 
 This has the potential to optimize the allocation of human hospital resources toward difficult cases, with the incorporated uncertainty representation allowing for transparent and safer patient-centered care.

\section{Conclusion}
In summary, we present a framework for uncertainty representation in healthcare, evaluated with a biomedical image segmentation task. The framework which can be generalized to other settings indicates the viability of uncertainty representations using MCU-Net combined with epistemic uncertainty to represent areas where the model is uncertain. Additionally, incorporating an uncertainty threshold would allow challenging cases with high uncertainty to be automatically referred to a medical-professional-in-the-loop. Moreover, we utilize uncertainty to address the unique aspect of healthcare by facilitating evaluation on a patient-by-patient basis rather than across the cohort. These promising initial results present opportunities for future research. Our framework could be applied on other models and application areas within healthcare both for classification and regression problems. This work is a step in the right direction towards uncertainty representations being leveraged to enable human-in-the loop healthcare systems.

\bibliographystyle{ACM-Reference-Format}
\bibliography{refs}

\end{document}